\documentclass{article}

\usepackage{microtype}
\usepackage{graphicx}
\usepackage{booktabs} % for professional tables
\usepackage{amsmath}

\usepackage{multirow}
\usepackage{array}
\usepackage{bm}
\usepackage{color}

\usepackage{makecell}
\usepackage{stackengine}
\usepackage{subcaption}
\usepackage{amssymb}
\usepackage{nicefrac}

\usepackage{hyperref}

\usepackage[accepted]{icml2019}

\icmltitlerunning{Self-Attention Generative Adversarial Networks}

\begin{document}

\twocolumn[
\icmltitle{Self-Attention Generative Adversarial Networks}

\icmlsetsymbol{equal}{*}

\begin{icmlauthorlist}
\icmlauthor{Han Zhang}{ru,gg}
\icmlauthor{Ian Goodfellow}{ga}
\icmlauthor{Dimitris Metaxas}{ru}
\icmlauthor{Augustus Odena}{gg}

\end{icmlauthorlist}

\icmlaffiliation{ru}{Department of Computer Science, Rutgers University}
\icmlaffiliation{gg}{Google Research, Brain Team}
\icmlaffiliation{ga}{Work done while at Google Research}

\icmlcorrespondingauthor{Han Zhang}{zhanghan@google.com}

\icmlkeywords{Machine Learning, ICML}

\vskip 0.3in
]



\printAffiliationsAndNotice{}  % leave blank if no need to mention equal contribution

\begin{abstract}
  In this paper, we propose the Self-Attention Generative Adversarial Network (SAGAN) which allows attention-driven, long-range dependency modeling for image generation tasks. Traditional convolutional GANs generate high-resolution details as a function of only spatially local points in lower-resolution feature maps. In SAGAN, details can be generated using cues from all feature locations. Moreover, the discriminator can check that highly detailed features in distant portions of the image are consistent with each other.
Furthermore, recent work has shown that generator conditioning affects GAN performance. Leveraging this insight, we apply spectral normalization to the GAN generator and find that this improves training dynamics. The proposed SAGAN performs better than prior work\footnote{\citet{BIGGAN}, which builds heavily on this work, has since improved those results substantially.}, boosting the best published Inception score from 36.8 to 52.52 and reducing Fr\'echet Inception distance from 27.62 to 18.65 on the challenging ImageNet dataset. Visualization of the attention layers shows that the generator leverages neighborhoods that correspond to object shapes rather than local regions of fixed shape.
\end{abstract}

%\vspace{-5pt}
\section{Introduction}\label{intro}

Image synthesis is an important problem in computer vision. There has been remarkable progress in this direction with the emergence of Generative Adversarial Networks (GANs)~\cite{goodfellow2014generative}, though many open problems remain ~\cite{OPENPROBLEMS}. GANs based on deep convolutional networks~\cite{Radford15,KarrasALL18,Han17stackgan2} have been especially successful. However, by carefully examining the generated samples from these models, we can observe that convolutional GANs~\cite{Odena2016,Miyato18a, Miyato18b} have much more difficulty in modeling some image classes than others when trained on multi-class datasets (\eg, ImageNet~\cite{ILSVRC15}). For example, while the state-of-the-art ImageNet GAN model~\cite{Miyato18b} excels at synthesizing image classes with few structural constraints (\eg, ocean, sky and landscape classes, which are distinguished more by texture than by geometry), it fails to capture geometric or structural patterns that occur consistently in some classes (for example, dogs are often drawn with realistic fur texture but without clearly defined separate feet). One possible explanation for this is that previous models rely heavily on convolution to model the dependencies across different image regions. Since the convolution operator has a local receptive field, long range dependencies can only be processed after passing through several convolutional layers. This could prevent learning about long-term dependencies for a variety of reasons: a small model may not be able to represent them, optimization algorithms may have trouble discovering parameter values that carefully coordinate multiple layers to capture these dependencies, and these parameterizations may be statistically brittle and prone to failure when applied to previously unseen inputs. Increasing the size of the convolution kernels can increase the representational capacity of the network but doing so also loses the computational and statistical efficiency obtained by using local convolutional structure. Self-attention~\cite{Cheng16,ParikhT0U16, Ashish17}, on the other hand, exhibits a better balance between the ability to model long-range dependencies and the computational and statistical efficiency. The self-attention module calculates response at a position as a weighted sum of the features at all positions, where the weights -- or attention vectors -- are calculated with only a small computational cost.

\begin{figure*}[tb]
    \centering
    \small
    \includegraphics[width=2\columnwidth]{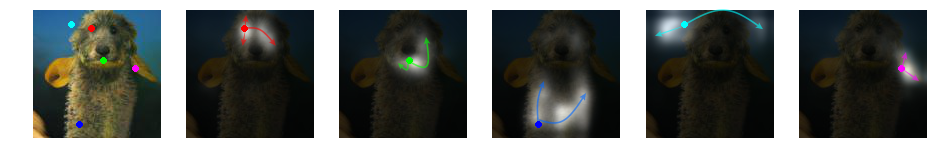}\\
    \vspace{-10pt}
    \includegraphics[width=2\columnwidth]{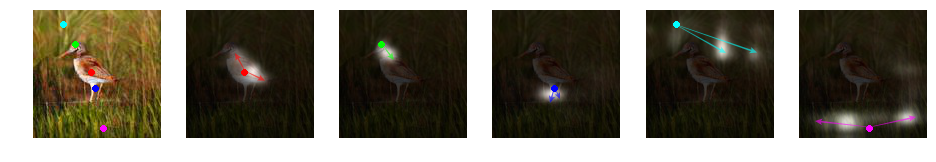}\\
    \vspace{-10pt}
    \caption{
    The proposed SAGAN generates images by leveraging complementary features in distant portions of the image rather than local regions of fixed shape to generate consistent objects/scenarios. 
    In each row, the first image shows five representative query locations with color coded dots. The other five images are attention maps for those query locations, with corresponding color coded arrows summarizing the most-attended regions. 
    }
    % \vspace{-10pt}
\label{fig:examples}
 \end{figure*}

In this work, we propose Self-Attention Generative Adversarial Networks (SAGANs), which introduce a self-attention mechanism into convolutional GANs. The self-attention module is complementary to convolutions and helps with  modeling long range, multi-level dependencies across image regions. Armed with self-attention, the generator can draw images in which fine details at every location are carefully coordinated with fine details in distant portions of the image. Moreover, the discriminator can also more accurately enforce complicated geometric constraints on the global image structure.

In addition to self-attention, we also incorporate recent insights relating network conditioning to GAN performance. The work by \cite{Odena18} showed that well-conditioned generators tend to perform better. We propose enforcing good conditioning of GAN generators using the spectral normalization technique that has previously been applied only to the discriminator \cite{Miyato18a}.

We have conducted extensive experiments on the ImageNet dataset to validate the effectiveness of the proposed self-attention mechanism and stabilization techniques. \textbf{SAGAN significantly outperforms prior work in image synthesis by boosting the best reported Inception score from  36.8 to 52.52 and reducing Fr\'echet Inception distance from 27.62 to 18.65.}
Visualization of the attention layers shows that the generator leverages neighborhoods that correspond to object shapes rather than local regions of fixed shape. Our code is available at \url{https://github.com/brain-research/self-attention-gan}.

\section{Related Work}\label{sec:sagan_relate}

\vspace{+10pt}
\textbf{Generative Adversarial Networks.} GANs have achieved great success in various image generation tasks, including image-to-image translation~\cite{pix2pix2017,cyclegan2017,Taigmaniclr17,LiuT16,Xue17segan,Taesung2019}, image super-resolution~\cite{Christian2016,Casper2016} and text-to-image synthesis~\cite{reed2016generative,reed2016learning,Han17, HongYCL18}.  Despite this success, the training of GANs is known to be unstable and sensitive to the choices of hyper-parameters.
Several works have attempted to stabilize the GAN training dynamics and improve the sample diversity by designing new network architectures~\cite{Radford15,Han17, KarrasALL18, Karras2019}, modifying the learning objectives and dynamics~\cite{Martin17WGAN,Salimans18,MetzICLR17, CheLJBL16, Zhao2016,Alexia2019}, adding regularization methods~\cite{GulrajaniAADC17,Miyato18a} and introducing heuristic tricks~\cite{Salimans2016, Odena2016, DRS}.
Recently, Miyato \etal~\cite{Miyato18a} proposed limiting the spectral norm of the weight matrices in the discriminator in order to constrain the Lipschitz constant of the discriminator function.
Combined with the projection-based discriminator~\cite{Miyato18b}, the spectrally normalized model greatly improves class-conditional image generation on ImageNet.

\vspace{+10pt}
\textbf{Attention Models.} Recently, attention mechanisms have become an integral part of models that must capture global dependencies~\cite{Dzmitry14,XuBKCCSZB15,YangHGDS16,GregorDGRW15, ChenMRA18}. 
In particular, self-attention~\cite{Cheng16, ParikhT0U16}, also called intra-attention, calculates the response at a position in a sequence by attending to all positions within the same sequence. 
Vaswani \etal~\cite{Ashish17} demonstrated that machine translation models could achieve state-of-the-art results by solely using a self-attention model. 
Parmar \etal~\cite{Parmar18} proposed an Image Transformer model to add self-attention into an autoregressive model for image generation. 
Wang \etal~\cite{Wang18} formalized self-attention as a non-local operation to model the spatial-temporal dependencies in video sequences. 
In spite of this progress, self-attention has not yet been explored in the context of GANs. 
(AttnGAN~\cite{Xu18} uses attention over word embeddings within an {\em input} sequence, but not self-attention over {\em internal model states}). 
SAGAN learns to efficiently find global, long-range dependencies within internal representations of images.

\begin{figure*}[tb]
%\begin{center}
\centering
\includegraphics[width=1.7\columnwidth]{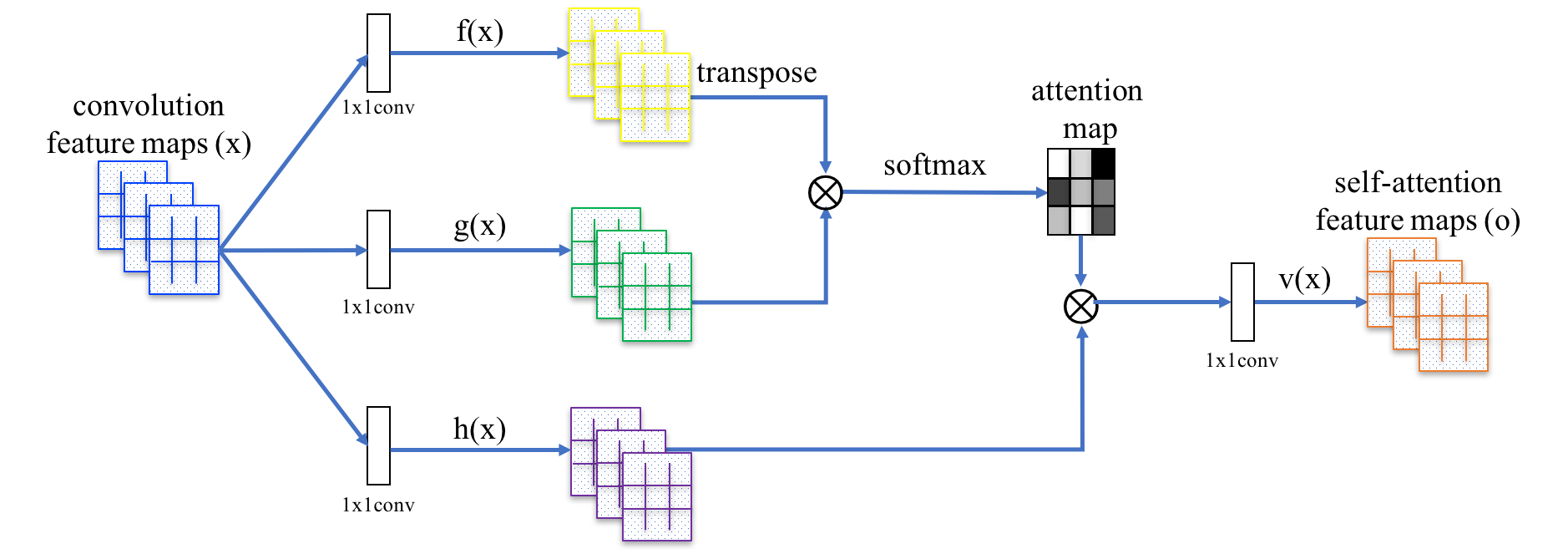}
%\end{center}
%  \vspace{-10pt}
   \caption{The proposed self-attention module for the SAGAN. The $\otimes$ denotes matrix multiplication. The softmax operation is performed on each row.}
% \vspace{-10pt}
  \label{fig:framework}
\end{figure*}

\section{Self-Attention Generative Adversarial Networks} \label{sec:sagan_sagan} 

Most GAN-based models~\cite{Radford15, Salimans2016, KarrasALL18} for image generation are built using convolutional layers. Convolution processes the information in a local neighborhood, thus using convolutional layers alone is computationally inefficient for modeling long-range dependencies in images.
In this section, we adapt the non-local model of \cite{Wang18} to introduce self-attention to the GAN framework, enabling both the generator and the discriminator to efficiently model relationships between widely separated spatial regions. We call the proposed method Self-Attention Generative Adversarial Networks (SAGAN) because of its self-attention module (see Figure~\ref{fig:framework}).

The image features from the previous hidden layer  $\bm{x} \in \mathbb{R}^{C \times N}$ are first transformed into two feature spaces $\bm{f}, \bm{g}$ to calculate the attention, where $\bm{f({x})}=\bm{W_f}\bm{x}, \; \bm{g({x})}=\bm{{W_g}{x}}$
\begin{equation}\label{eq:attention}
%\begin{aligned}
\beta_{j, i} = \frac{\exp (s_{ij})}{\sum_{i=1}^{N} \exp(s_{ij})},\; \text{where} \; s_{ij}= \bm{f}(\bm{x_i})^T \bm{g}(\bm{x_j}),
%\beta_{j, i} = \frac{\exp (\bm{f(x_i)}^T \bm{g(x_j)})}{\sum_{i=1}^{M\times N} %\exp(\bm{f(x_i)}^T \bm{g(x_j)})}
%\end{aligned}
\end{equation}
and $\beta_{j, i}$ indicates the extent to which the model attends to the $i^{th}$ location when synthesizing the $j^{th}$ region. Here, $C$ is the number of channels and $N$ is the number of feature locations of features from the previous hidden layer.
The output of the attention layer is $\bm{o} = (\bm{o_1}, \bm{o_2}, ..., \bm{o_j}, ..., \bm{o_N}) \in \mathbb{R}^{C \times N}$, where,
\begin{equation}\label{eq:attention_out}
\begin{split}
\bm{o_j} = \bm{v}\left(\sum_{i=1}^{N} \beta_{j, i} \bm{h}(\bm{x_i})\right), \; \bm{h}(\bm{x_i})=\bm{{W_h}{x_i}}, \;  \bm{v}(\bm{x_i})=\bm{{W_v}{x_i}}.
\end{split}
\end{equation}
In the above formulation, $\bm{W_g} \in \mathbb{R}^{\bar{C} \times C}$, $\bm{W_f} \in \mathbb{R}^{\bar{C} \times C}$, $\bm{W_h} \in \mathbb{R}^{\bar{C} \times C}$, and $\bm{W_v} \in \mathbb{R}^{C \times \bar{C}}$ are the learned weight matrices, which are implemented as  1$\times$1 convolutions. 
Since We did not notice any significant performance decrease when reducing the channel number of $\bar{C}$ to be $\nicefrac{C}{k}$, where $k=1, 2, 4, 8$ after few training epochs on ImageNet. For memory efficiency, we choose $k=8$ (\ie, $\bar{C} = \nicefrac{C}{8}$) in all our experiments.

In addition, we further multiply the output of the attention layer by a scale parameter and add back the input feature map. Therefore, the final output is given by, 
\begin{equation}\label{eq:attention_output}
\bm{y_i} = \gamma \bm{o_i} + \bm{x_i}, 
\end{equation}

where $\gamma$ is a learnable scalar and it is initialized as 0.
Introducing the learnable $\gamma$ allows the network to first rely on the cues in the local neighborhood -- since this is easier -- and then gradually learn to assign more weight to the non-local evidence.
The intuition for why we do this is straightforward: we want to learn the easy task first and then progressively increase the complexity of the task.
In the SAGAN, the proposed attention module has been applied to both the generator and the discriminator, which are trained in an alternating fashion by minimizing the hinge version of the adversarial loss~\cite{lim2017,Tran2017,Miyato18a},
%\vspace{-5pt}
\begin{equation}\label{eq:GAN_Loss}
\begin{aligned}
L_D = \; & -\mathbb{E}_{(x, y) \sim {p_{data}}} [\min(0, -1 + D(x, y))] \\ &- \mathbb{E}_{z \sim {p_{z}}, y \sim {p_{data}}} [\min(0, -1-D(G(z), y)) ], \\
L_G = \;  &-\mathbb{E}_{z \sim {p_{z}}, y \sim {p_{data}}} {D(G(z), y)},
\end{aligned}
\end{equation}
%\vspace{-5pt}

\section{Techniques to Stabilize the Training of GANs}

We also investigate two techniques to stabilize the training of GANs on challenging datasets.
First, we use spectral normalization~\cite{Miyato18a} in the generator as well as in the discriminator.
Second, we confirm that the two-timescale update rule (TTUR)~\cite{HeuselRUNH17} is effective, and we advocate using it specifically to address slow learning in regularized discriminators.

\subsection{Spectral normalization for both generator and discriminator}

Miyato \etal~\cite{Miyato18a} originally proposed stabilizing the training of GANs by applying spectral normalization to the discriminator network.
Doing so constrains the Lipschitz constant of the discriminator by restricting the spectral norm of each layer.
Compared to other normalization techniques, spectral normalization does not require extra hyper-parameter tuning (setting the spectral norm of all weight layers to $1$ consistently performs well in practice).
Moreover, the computational cost is also relatively small.

We argue that the generator can also benefit from spectral normalization, based on recent evidence that the conditioning of the generator is an important causal factor in GANs' performance \cite{Odena18}.
Spectral normalization in the generator can prevent the escalation of parameter magnitudes and avoid unusual gradients.
We find empirically that spectral normalization of both generator and discriminator makes it possible to use fewer discriminator updates per generator update, thus significantly reducing the computational cost of training. The approach also shows more stable training behavior.

\subsection{Imbalanced learning rate for generator and discriminator updates}

In previous work, regularization of the discriminator~\cite{Miyato18a, GulrajaniAADC17} often slows down the GANs' learning process.
In practice, methods using regularized discriminators typically require multiple (\eg, 5) discriminator update steps per generator update step during training.
Independently, Heusel \etal~\cite{HeuselRUNH17} have advocated using separate learning rates (TTUR) for the generator and the discriminator.
We propose using TTUR specifically to compensate for the problem of slow learning in a regularized discriminator, making it possible to use fewer discriminator steps per generator step.
Using this approach, we are able to produce better results given the same wall-clock time.

\section{Experiments}\label{sec:sagan_exp}

To evaluate the proposed methods, we conducted extensive experiments on the LSVRC2012 (ImageNet) dataset~\cite{ILSVRC15}. First, in Section~\ref{sec:sagan_stable}, we present experiments designed to evaluate the effectiveness of the two proposed techniques for stabilizing GANs' training. Next, the proposed self-attention mechanism is investigated in Section~\ref{sec:sagan_component}. Finally, our SAGAN is compared with state-of-the-art methods~\cite{Odena2016,Miyato18b} on the image generation task in Section~\ref{sec:sagan_compare}.
Models were trained for roughly 2 weeks on 4 GPUs each, using sychronous SGD (as there are well known difficulties with asynchronous SGD - see e.g. \cite{FASGD}).

\textbf{Evaluation metrics. }
{
  We choose the Inception score (IS)~\cite{Salimans2016} and the Fr\'echet Inception distance (FID)~\cite{HeuselRUNH17} for quantitative evaluation.
  Though alternatives exist ~\cite{HYPE,GEOMETRYSCORE,SKILLRATING}, they are not widely used.
  The Inception score~\cite{Salimans2016} computes the KL divergence between the conditional class distribution and the marginal class distribution. Higher Inception score indicates better image quality. We include the Inception score because it is widely used and thus makes it possible to compare our results to previous work. However, it is important to understand that Inception score has serious limitations---it is intended primarily to ensure that the model generates  samples that can be confidently recognized as belonging to a specific class, and that the model generates samples from many classes, not necessarily to assess realism of details or intra-class diversity.
 FID is a more principled and comprehensive metric, and has been shown to be more consistent with human evaluation in assessing the realism and variation of the generated samples~\cite{HeuselRUNH17}. FID calculates the Wasserstein-2 distance between the generated images and the real images in the feature space of an Inception-v3 network. 
 Besides the FID calculated over the whole data distribution (\ie., all 1000 classes of images in ImageNet), we also compute FID between the generated images and dataset images within each class (called intra FID~\cite{Miyato18b}). Lower FID and intra FID values mean closer distances between synthetic and real data distributions. In all our experiments, 50k samples are randomly generated for each model to compute the Inception score, FID and intra FID. 
}

%
%%
% \begin{figure*}[tb]
%     \centering
%     \small
%     \begin{tabular}{c@{\hspace{2mm}}c@{\hspace{2mm}}c@{\hspace{2mm}}c}
%     \stackunder[3pt]{\makecell[l]{
%         \includegraphics[width=0.315\columnwidth]{sagan/base1-fid-new}}}{}\vspace{+2pt}&
%     \stackunder[3pt]{\makecell[l]{
%         \includegraphics[width=0.315\columnwidth]{sagan/gSN-fid-new}}}{}\vspace{+2pt}&
%     \stackunder[3pt]{\makecell[l]{
%         \includegraphics[width=0.315\columnwidth]{sagan/gSN-LR-fid-new}}}{}\vspace{+2pt}\\
%     \stackunder[3pt]{\makecell[l]{
%         \includegraphics[width=0.315\columnwidth]{sagan/base1-is-new.png}}}{}\vspace{+2pt}&
%     \stackunder[3pt]{\makecell[l]{
%         \includegraphics[width=0.315\columnwidth]{sagan/gSN-is-new}}}{}\vspace{+2pt}&
%     \stackunder[3pt]{\makecell[l]{
%         \includegraphics[width=0.315\columnwidth]{sagan/gSN-LR-is-new}}}{}\vspace{+2pt}\\
%     \end{tabular}
%     % \vspace{-8pt}
%     \caption{Training curves for the baseline model and our models with the proposed stabilization techniques, ``SN on $G$/$D$'' and two-timescale learning rates (TTUR). All models are trained with 1:1 balanced updates for $G$ and $D$.}
% \label{fig:stable}
%     % \vspace{-5pt}
%  \end{figure*}
%%

\begin{figure*}[tb]
    \centering
    \small
    \begin{tabular}{c@{\hspace{2mm}}c@{\hspace{2mm}}c@{\hspace{2mm}}c}
    \stackunder[3pt]{\makecell[l]{
        \includegraphics[width=0.28\textwidth]{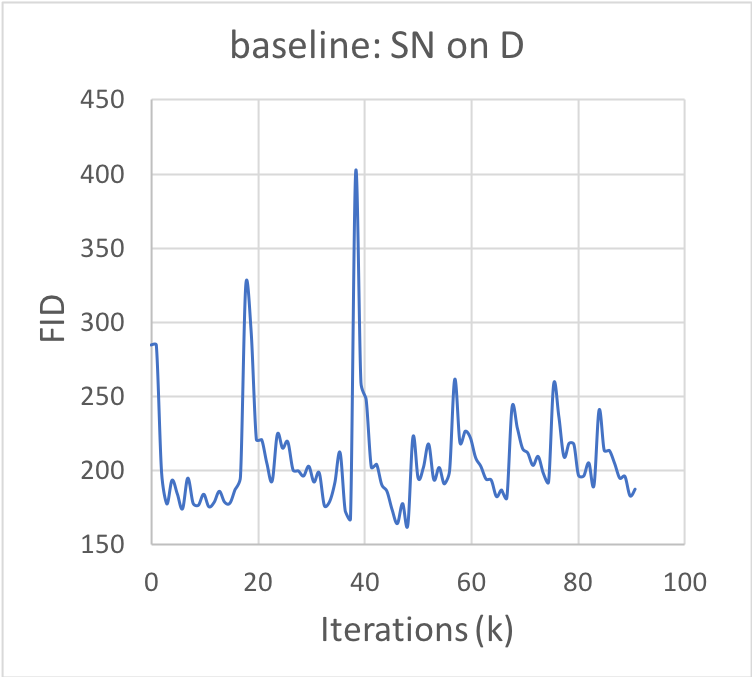}}}{}\vspace{+2pt}&
    \stackunder[3pt]{\makecell[l]{
        \includegraphics[width=0.28\textwidth]{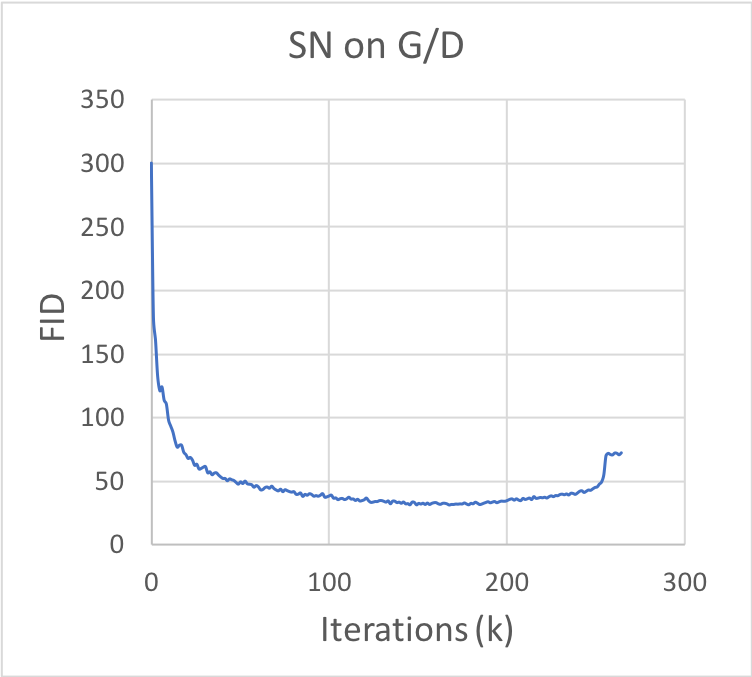}}}{}\vspace{+2pt}&
    \stackunder[3pt]{\makecell[l]{
        \includegraphics[width=0.28\textwidth]{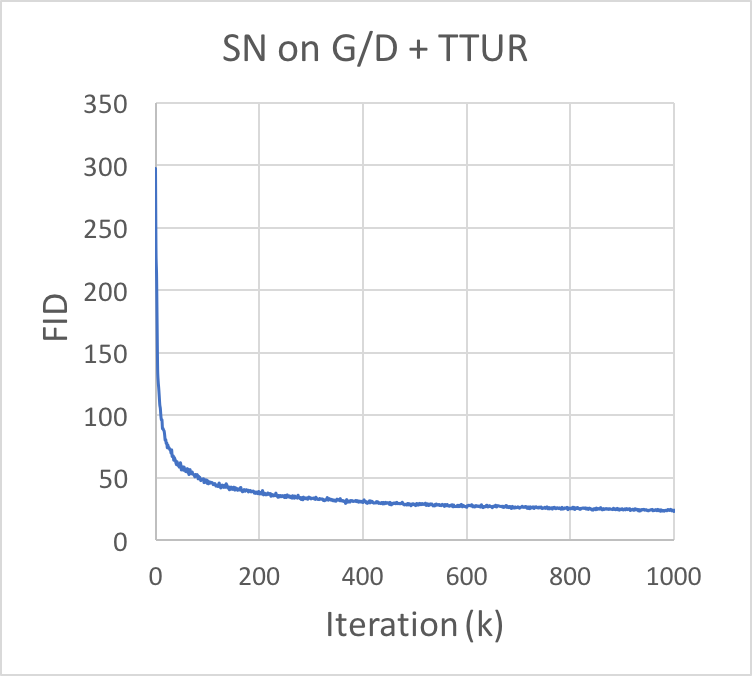}}}{}\vspace{+2pt}\\
    \stackunder[3pt]{\makecell[l]{
        \includegraphics[width=0.28\textwidth]{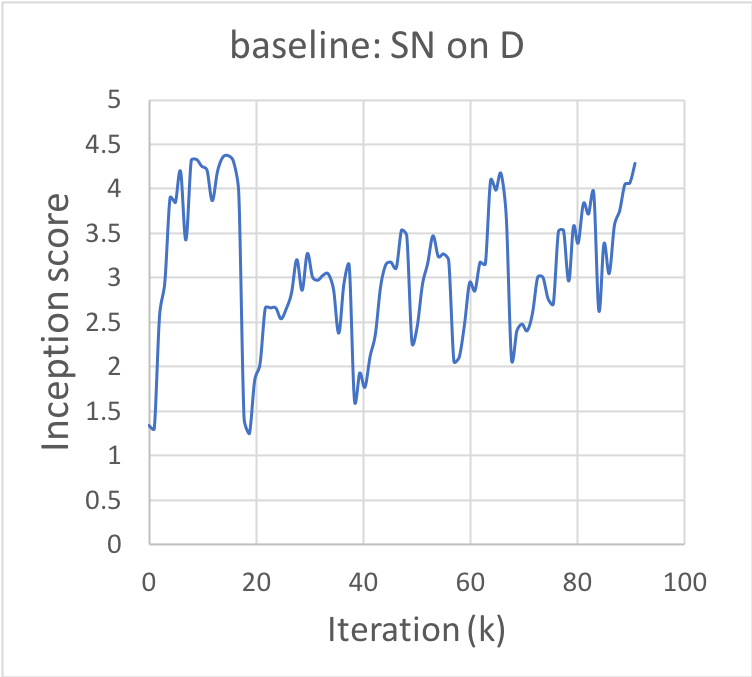}}}{}\vspace{+2pt}&
    \stackunder[3pt]{\makecell[l]{
        \includegraphics[width=0.28\textwidth]{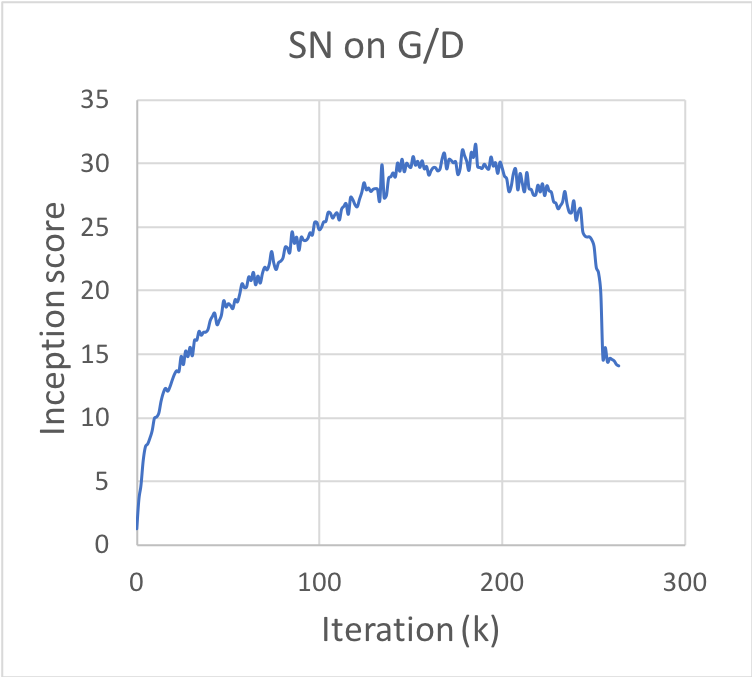}}}{}\vspace{+2pt}&
    \stackunder[3pt]{\makecell[l]{
        \includegraphics[width=0.28\textwidth]{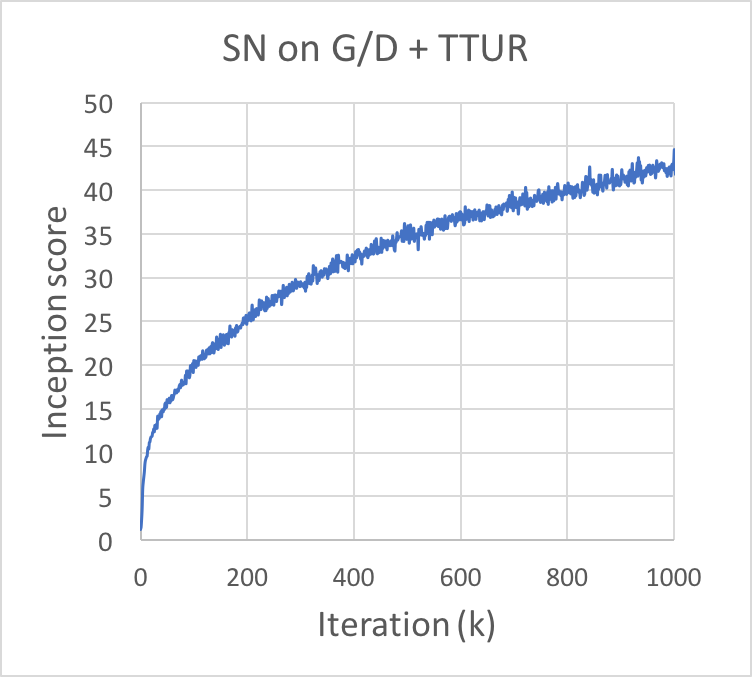}}}{}\vspace{+2pt}\\
    \end{tabular}
    % \vspace{-8pt}
    \caption{Training curves for the baseline model and our models with the proposed stabilization techniques, ``SN on $G$/$D$'' and two-timescale learning rates (TTUR). All models are trained with 1:1 balanced updates for $G$ and $D$.}
\label{fig:stable}
    % \vspace{-5pt}
 \end{figure*}

\begin{figure*}[tb]
    \centering
    \small
    \begin{tabular}{c@{\hspace{2mm}}c@{\hspace{2mm}}c@{\hspace{2mm}}c}
    \stackunder[3pt]{\makecell[l]{
        \includegraphics[width=0.22\textwidth]{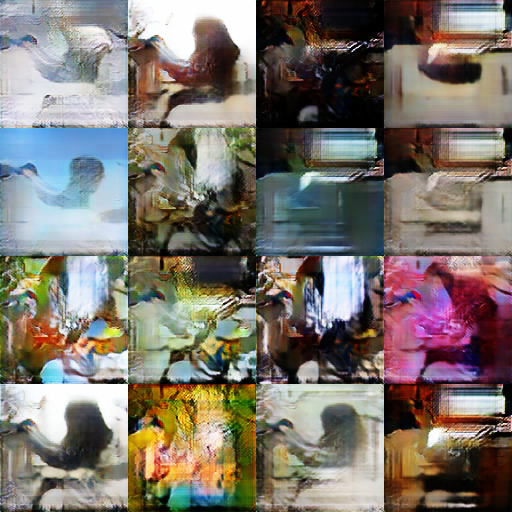}}}{\shortstack{Baseline: SN on D \\(10k,~FID=181.84)}}\vspace{+2pt}&
    \stackunder[3pt]{\makecell[l]{
        \includegraphics[width=0.22\textwidth]{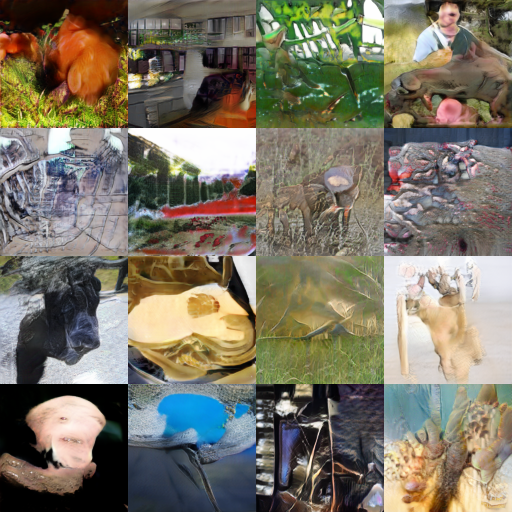}}}{\shortstack{SN on $G$/$D$ \\(10k,~FID=93.52)}}\vspace{+2pt}&
    \stackunder[3pt]{\makecell[l]{
        \includegraphics[width=0.22\textwidth]{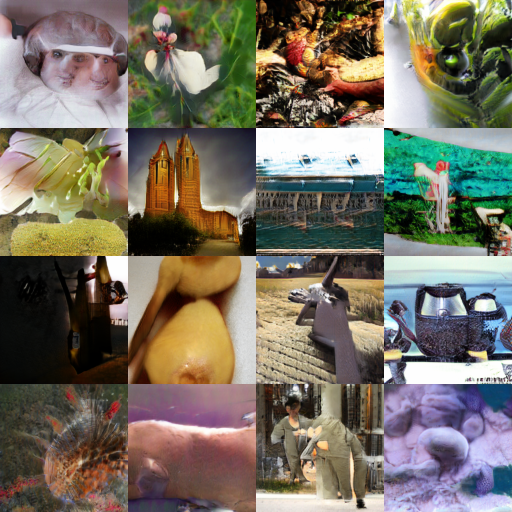}}}{\shortstack{SN on $G$/$D$ \\(160k,~FID=33.39)}}\vspace{+2pt}&
    \stackunder[3pt]{\makecell[l]{
        \includegraphics[width=0.22\textwidth]{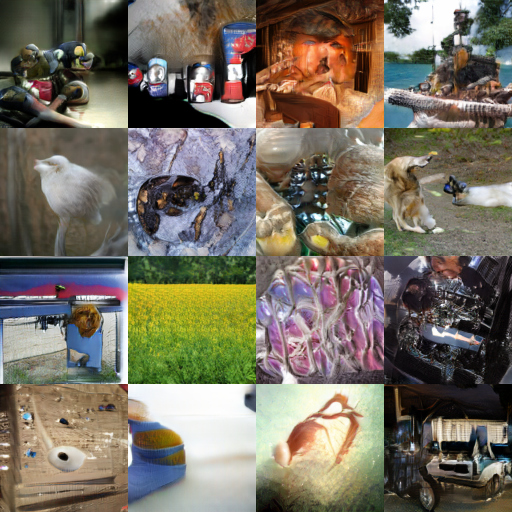}}}{\shortstack{SN on $G$/$D$ \\(260k,~FID=72.41)}}\vspace{+2pt}\\
    \stackunder[3pt]{\makecell[l]{
        \includegraphics[width=0.22\textwidth]{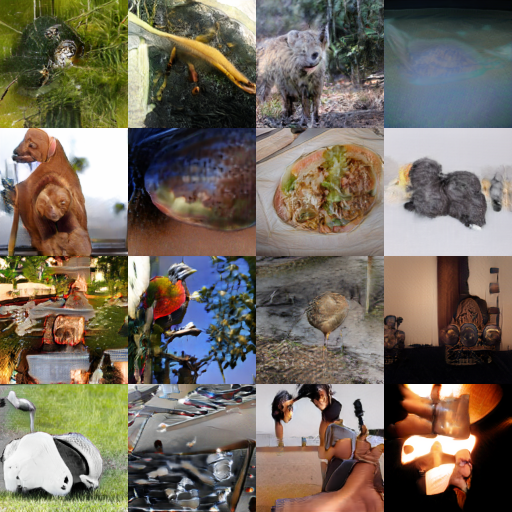}}}{\shortstack{SN on $G$/$D$+TTUR\\ (10k,~FID=99.04)}}\vspace{+2pt}&
    \stackunder[3pt]{\makecell[l]{
        \includegraphics[width=0.22\textwidth]{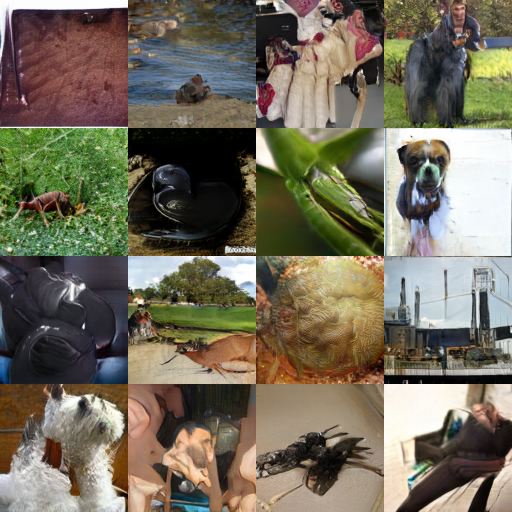}}}{\shortstack{SN on $G$/$D$+TTUR\\ (160k,~FID=40.96)}}\vspace{+2pt}&
    \stackunder[3pt]{\makecell[l]{
        \includegraphics[width=0.22\textwidth]{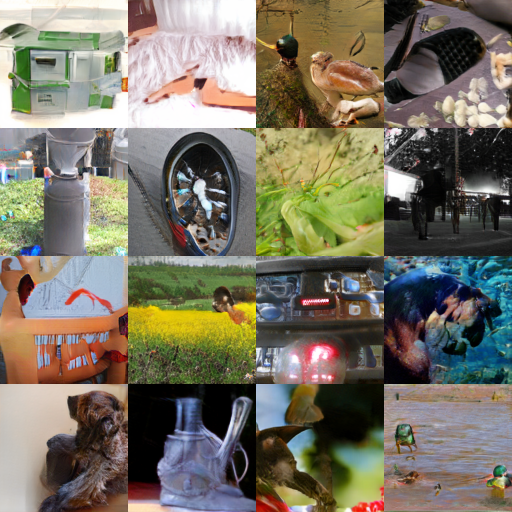}}}{\shortstack{SN on $G$/$D$+TTUR\\ (260k,~FID=34.62)}}\vspace{+2pt}&
    \stackunder[3pt]{\makecell[l]{
        \includegraphics[width=0.22\textwidth]{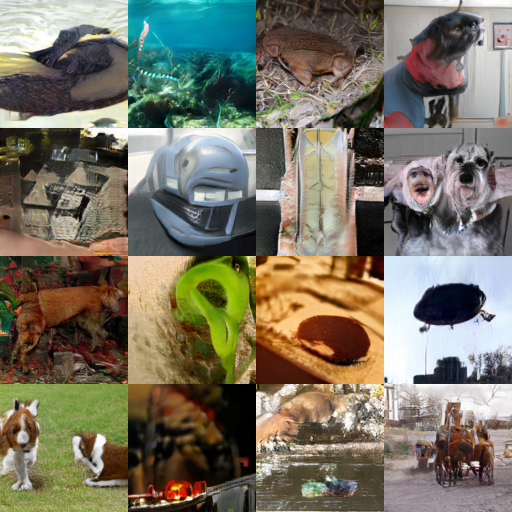}}}{\shortstack{SN on $G$/$D$+TTUR\\ (1M,~FID=22.96)}}\vspace{+2pt}\\
    \end{tabular}
    \vspace{-5pt}
    \caption{128$\times$128 examples randomly generated by the baseline model and our models ``SN on $G$/$D$'' and ``SN on $G$/$D$+TTUR''.}
\label{fig:stable_examples}
    % \vspace{-10pt}
 \end{figure*}

\textbf{Network structures and implementation details. }
{
All the SAGAN models we train are designed to generate 128$\times$128 images. 
% The detailed network structure is included in Appendix~A. 
By default, spectral normalization~\cite{Miyato18a} is used for the layers in both the generator and the discriminator. Similar to \cite{Miyato18b}, SAGAN uses conditional batch normalization in the generator and projection in the discriminator. For all models, we use the Adam optimizer ~\cite{KingmaB14} with $\beta_1 = 0$ and $\beta_2 = 0.9$ for training.
By default, the learning rate for the discriminator is 0.0004 and the learning rate for the generator is 0.0001. 
}

\subsection{Evaluating the proposed stabilization techniques} \label{sec:sagan_stable}

In this section, experiments are conducted to evaluate the effectiveness of the proposed stabilization techniques, \ie, applying spectral normalization (SN) to the generator and utilizing imbalanced learning rates (TTUR). In Figure~\ref{fig:stable}, our models ``SN on $G$/$D$'' and ``SN on $G$/$D$+TTUR'' are compared with a baseline model, which is implemented based on the state-of-the-art image generation method~\cite{Miyato18a}. In this baseline model, SN is only utilized in the discriminator. When we train it with 1:1 balanced updates for the discriminator ($D$) and the generator ($G$), the training becomes very unstable, as shown in the leftmost sub-figures of Figure~\ref{fig:stable}. It exhibits mode collapse very early in training. For example, the top-left sub-figure of Figure~\ref{fig:stable_examples} illustrates some images randomly generated by the baseline model at the 10k-th iteration. Although in the the original paper~\cite{Miyato18a} this unstable training behavior is greatly mitigated by using 5:1 imbalanced updates for $D$ and $G$, the ability to be stably trained with 1:1 balanced updates is desirable for improving the convergence speed of the model. Thus, using our proposed techniques means that the model can produce better results given the same wall-clock time. Given this, there is no need to search for a suitable update ratio for the generator and discriminator. As shown in the middle sub-figures of Figure~\ref{fig:stable}, adding SN to both the generator and the discriminator greatly stabilized our model ``SN on $G$/$D$'', even when it was trained with 1:1 balanced updates. However, the quality of samples does not improve monotonically during training. For example, the image quality as measured by FID and IS is starting to drop at the 260k-th iteration. Example images randomly generated by this model at different iterations can be found in Figure~\ref{fig:stable_examples}. When we also apply the imbalanced learning rates to train the discriminator and the generator, the quality of images generated by our model ``SN on $G$/$D$+TTUR'' improves monotonically during the whole training process. As shown in Figure~\ref{fig:stable} and Figure~\ref{fig:stable_examples}, we do not observe any significant decrease in sample quality or in the FID or the Inception score during one million training iterations. Thus, both quantitative results and qualitative results demonstrate the effectiveness of the proposed stabilization techniques for GANs' training. They also demonstrate that the effect of the two techniques is at least partly additive. In the rest of experiments, all models use spectral normalization for both the generator and discriminator and use the imbalanced learning rates to train the generator and the discriminator with 1:1 updates.

\begin{table*}[bt]
\begin{center}
\small
% \scriptsize
\begin{tabular}{|c|c|c|c|c|c|c|c|c|c|}
\hline
\multirow{2}{2.5em}{Model} & \multirow{2}{3.2em}{\shortstack{no\\ attention}} & \multicolumn{4}{c|}{SAGAN} & \multicolumn{4}{c|}{Residual} \\
\cline{3-10}
& &$feat_{8}$ &$feat_{16}$ &$feat_{32}$ &$feat_{64}$ &$feat_{8}$ &$feat_{16}$ &$feat_{32}$ &$feat_{64}$ \\
\hline
FID &22.96 &22.98 &22.14  &\bf18.28 &18.65 &42.13 &22.40 &27.33 &28.82\\
\hline
IS  &42.87 &43.15 & 45.94 &51.43 &\bf52.52 &23.17 &44.49 &38.50 &38.96\\
\hline
\end{tabular}
\end{center}
% \vspace{-8pt}
\caption{Comparison of Self-Attention and Residual block on GANs. These blocks are added into different layers of the network.
  All models have been trained for one million iterations, and the best Inception scores (IS) and  Fr\'echet Inception distance (FID) are reported. $feat_{k}$ means adding self-attention to the k$\times$k feature maps. }
% \vspace{-10pt}
\label{tab:cmp_attn} 
\end{table*}
%\FloatBarrier
%
%
 
%
%%
\begin{figure*}[tb]
    \centering
    \small
    \begin{tabular}{@{\hspace{0mm}}|c@{\hspace{1mm}}|@{\hspace{1mm}}c|@{\hspace{0mm}}}
    \hline
    \hspace{-9pt}\stackunder[0pt]{\makecell[l]{
        \includegraphics[width=0.48\textwidth]{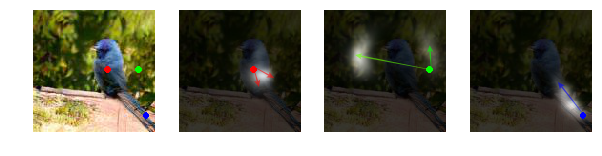}}}{}&
    \hspace{-9pt}\stackunder[0pt]{\makecell[l]{
        \includegraphics[width=0.48\textwidth]{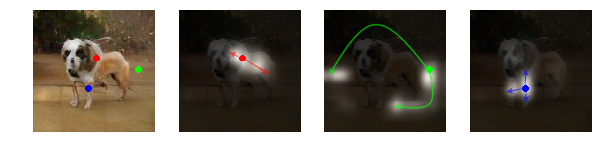}}}{}\\
    \hline
    \hspace{-9pt}\stackunder[0pt]{\makecell[l]{
        \includegraphics[width=0.48\textwidth]{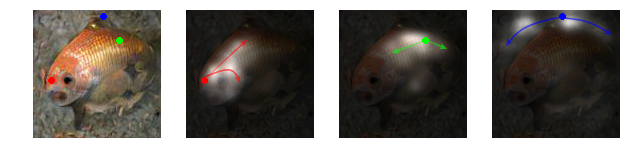}}}{}&
    \hspace{-9pt}\stackunder[0pt]{\makecell[l]{
        \includegraphics[width=0.48\textwidth]{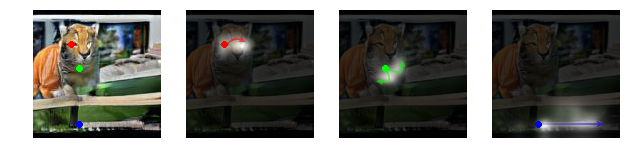}}}{}\\
    \hline
    \hspace{-9pt}\stackunder[0pt]{\makecell[l]{
        \includegraphics[width=0.48\textwidth]{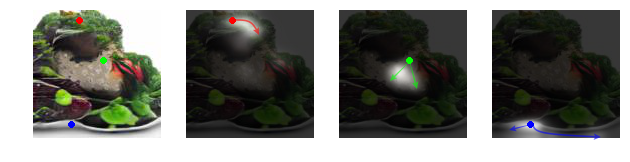}}}{}&
    \hspace{-9pt}\stackunder[0pt]{\makecell[l]{
        \includegraphics[width=0.48\textwidth]{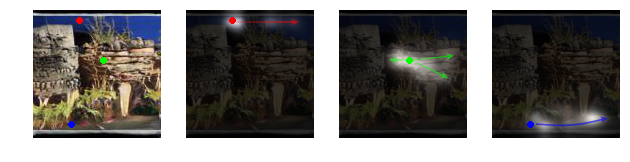}}}{}\\
    \hline
    \hspace{-9pt}\stackunder[0pt]{\makecell[l]{
        \includegraphics[width=0.48\textwidth]{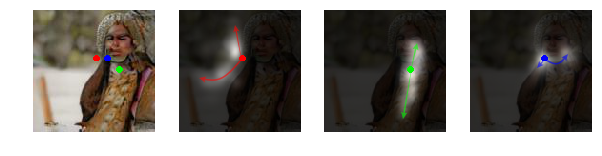}}}{}&
    \hspace{-9pt}\stackunder[0pt]{\makecell[l]{
        \includegraphics[width=0.48\textwidth]{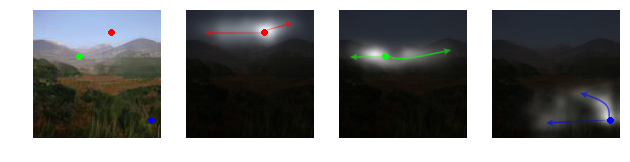}}}{}\\
    \hline
    \end{tabular}
    % \vspace{-5pt}
    \caption{Visualization of attention maps. 
    These images were generated by SAGAN.  
    We visualize the attention maps of the last generator layer that used attention, since this layer is the closest to the output pixels and is the most straightforward to project into pixel space and interpret. 
    In each cell, the first image shows three representative query locations with color coded dots. 
    The other three images are attention maps for those query locations, with corresponding color coded arrows summarizing the most-attended regions. 
    We observe that the network learns to allocate attention according to similarity of color and texture, rather than just spatial adjacency (see the top-left cell). 
    % For example, in the top-left cell, the red point attends mostly to the body of the bird around it, however, the green point learns to attend to other side of the image. 
    % In this way, the image has a consistent background (\ie, trees from the left to the right though they are separated by the bird). 
    % Similarly, the blue point allocates the attention to the whole tail of the bird to make the generated part coherent. 
    % Those long-range dependencies could not be captured by convolutions with local receptive fields. 
    We also find that although some query points are quite close in spatial location, their attention maps can be very different, as shown in the bottom-left cell. 
    % The red point attends mostly to the background regions, whereas the blue point, though adjacent to red point, puts most of the attention on the foreground object. 
    % This also reduces the chance for the local errors to propagate, since the adjacent position has the freedom to choose to attend to other distant locations. 
    % These observations further demonstrate that self-attention is complementary to convolutions for image generation in GANs. 
    As shown in the top-right cell, SAGAN is able to draw dogs with clearly separated legs. 
    The blue query point shows that attention helps to get the structure of the joint area correct. 
    See the text for more discussion about the properties of learned attention maps. 
    }
\label{fig:attention}
    \vspace{-8pt}
 \end{figure*}

\subsection{Self-attention mechanism.} \label{sec:sagan_component}

To explore the effect of the proposed self-attention mechanism, we build several SAGAN models by adding the self-attention mechanism to different stages of the generator and the discriminator. As shown in Table~\ref{tab:cmp_attn}, the SAGAN models with the self-attention mechanism at the middle-to-high level feature maps (\eg, $feat_{32}$ and $feat_{64}$) achieve better performance than the models with the self-attention mechanism at the low level feature maps (\eg, $feat_{8}$ and $feat_{16}$). For example, the FID of the model ``SAGAN, $feat_{8}$'' is improved from 22.98 to 18.28 by ``SAGAN, $feat_{32}$''. The reason is that self-attention receives more evidence and enjoys more freedom to choose conditions with larger feature maps (\ie, it is complementary to convolution for large feature maps), however, it plays a similar role as the local convolution when modeling dependencies for small (\eg, 8$\times$8) feature maps. It demonstrates that the attention mechanism gives more power to both the generator and the discriminator to directly model the long-range dependencies in the feature maps. In addition, the comparison of our SAGAN and the baseline model without attention (2nd column of Table~\ref{tab:cmp_attn}) further shows the effectiveness of the proposed self-attention mechanism.

Compared with residual blocks with the same number of parameters, the self-attention blocks also achieve better results. For example, the training is not stable when we replace the self-attention block with the residual block in 8$\times$8 feature maps, which leads to a significant decrease in performance (\eg, FID increases from 22.98 to 42.13). Even for the cases when the training goes smoothly, replacing the self-attention block with the residual block still leads to worse results in terms of FID and Inception score. (\eg, FID 18.28 vs 27.33 in feature map 32 $\times$ 32). This comparison demonstrates that the performance improvement given by using SAGAN is not simply due to an increase in model depth and capacity.

To better understand what has been learned during the generation process, we visualize the attention weights of the generator in SAGAN for different images. Some sample images with attention are shown in Figure~\ref{fig:attention} and Figure~\ref{fig:examples}. 
See the caption of Figure~\ref{fig:attention} for descriptions of some of the properties of learned attention maps.

%\clearpage
% \FloatBarrier
\subsection{Comparison with the state-of-the-art} \label{sec:sagan_compare}

Our SAGAN is also compared with the state-of-the-art GAN models~\cite{Odena2016,Miyato18b} for class conditional image generation on ImageNet. As shown in Table~\ref{tab:compare_others}, our proposed SAGAN achieves the best Inception score, intra FID and FID. The proposed SAGAN significantly improves the best published Inception score from 36.8 to 52.52. The lower FID (18.65) and intra FID (83.7) achieved by the SAGAN also indicates that the SAGAN can better approximate the original image distribution by using the self-attention module to model the long-range dependencies between image regions. 

\begin{figure*}[tb]
    \centering
    \small
    \begin{tabular}{cc}
     \shortstack{goldfish \\ ({\bf44.4}, 58.1)} &\includegraphics[width=0.85\textwidth]{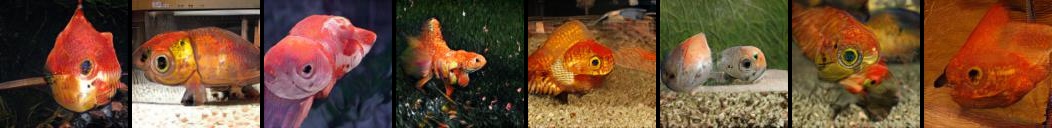}\\
    \shortstack{indigo \\bunting \\ ({\bf53.0}, 66.8)} &\includegraphics[width=0.85\textwidth]{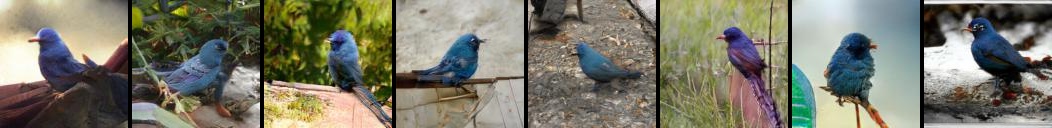}\\
    \shortstack{redshank \\ ({\bf48.9}, 60.1)} &\includegraphics[width=0.85\textwidth]{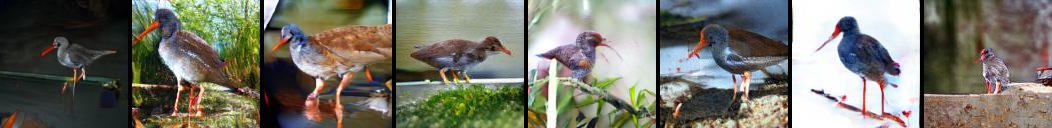}\\
    \shortstack{saint \\bernard \\({\bf35.7}, 55.3)} &\includegraphics[width=0.85\textwidth]{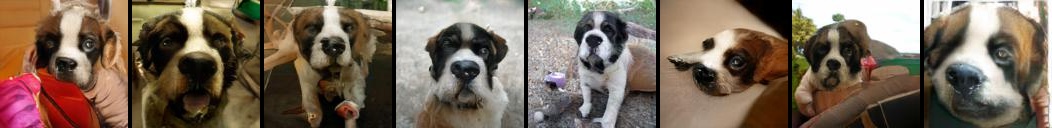}\\
    \shortstack{tiger \\cat \\({\bf88.1}, 90.2)} &\includegraphics[width=0.85\textwidth]{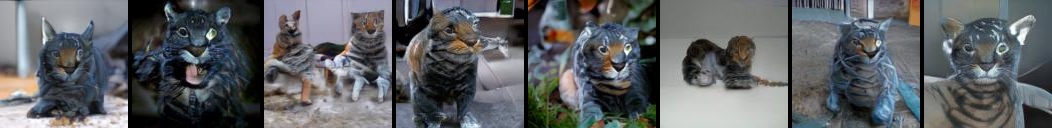}\\
    \shortstack{stone \\wall \\(57.5, {\bf49.3})} &\includegraphics[width=0.85\textwidth]{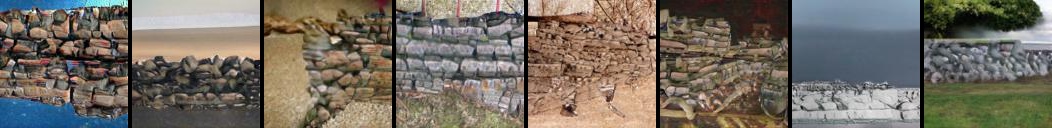}\\
    % \shortstack{broccoli \\(106.6, {\bf58.0})} &\includegraphics[width=0.865\columnwidth]{sagan/exp-937-broccoli.jpg}\\
    \shortstack{geyser \\(21.6, {\bf19.5})} &\includegraphics[width=0.85\textwidth]{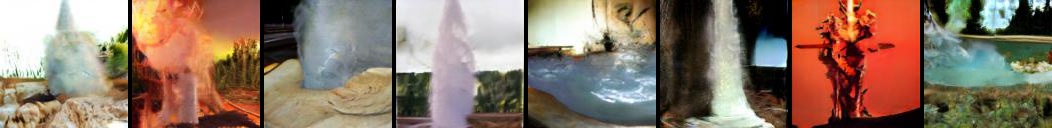}\\
    \shortstack{valley \\(39.7, {\bf26.0})} &\includegraphics[width=0.85\textwidth]{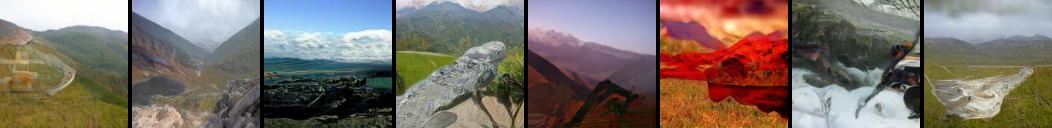}\\
    % \shortstack{rapeseed \\(104.7, {\bf17.9}))} &\includegraphics[width=0.865\columnwidth]{sagan/exp-984-rapeseed.jpg}\\
    \shortstack{coral \\fungus \\ (38.0, {\bf37.2})} &\includegraphics[width=0.85\textwidth]{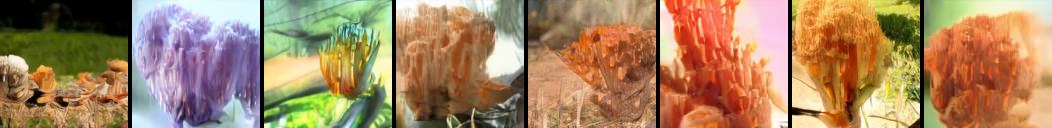}\\
    \end{tabular}
    %\vspace{-5pt}
    % \caption{Intra FID scores of (our SAGAN, the state-of-the-art method~\cite{Miyato18b}) (Left) and samples generated by our SAGAN (Right) for some representative classes of ImageNet. 
    % Each row shows results from one class. 
    % }
    \caption{128x128 example images generated by SAGAN for different classes. Each row shows examples from one class. In the leftmost column, the intra FID of our SAGAN (\emph{left}) and the state-of-the-art method~\cite{Miyato18b}) (\emph{right}) are listed. }
%\vspace{-10pt}
\label{fig:fish}
 \end{figure*}

Figure~\ref{fig:fish} shows some comparison results and generated-images for representative classes of ImageNet.  We observe that our SAGAN achieves much better performance (\ie, lower intra FID) than the state-of-the-art GAN model~\cite{Miyato18b} for synthesizing image classes with complex geometric or structural patterns, such as goldfish and Saint Bernard. For classes with few structural constraints (\eg, valley, stone wall and coral fungus, which are distinguished more by texture than by geometry), our SAGAN shows less superiority compared with the baseline model~\cite{Miyato18b}. Again, the reason is that the self-attention in SAGAN is complementary to the convolution for capturing long-range, global-level dependencies occurring consistently in geometric or structural patterns, but plays a similar role as the local convolution when modeling dependencies for simple texture.  
\begin{table*}[bt]
\begin{center}
\normalfont
% \scriptsize
\begin{tabular}{|c|c|c|c|c|}
\hline
Model                & Inception Score   &Intra FID   & FID \\
\hline
AC-GAN~\cite{Odena2016}              &28.5  &260.0       &/\ \\
\hline
SNGAN-projection~\cite{Miyato18b} &36.8  &92.4        & 27.62$^*$ \\
\hline
SAGAN                             &\bf52.52 &\bf 83.7 &\bf18.65 \\
\hline
\end{tabular}
\end{center}
%\vspace{-8pt}
    \caption{Comparison of the proposed SAGAN with state-of-the-art GAN models~\cite{Odena2016,Miyato18b} for class conditional image generation on ImageNet. FID of SNGAN-projection is calculated from officially released weights.}
%\vspace{-10pt}
\label{tab:compare_others} 
\end{table*}

\section{Conclusion}\label{sec:sagan_conclude}
In this paper, we proposed Self-Attention Generative Adversarial Networks (SAGANs), which incorporate a self-attention mechanism into the GAN framework.
The self-attention module is effective in modeling long-range dependencies.
In addition, we show that spectral normalization applied to the generator stabilizes GAN training and that TTUR speeds up training of regularized discriminators.
SAGAN achieves the state-of-the-art performance on class-conditional image generation on ImageNet.
\clearpage

\section*{Acknowledgments}
We thank Surya Bhupatiraju for feedback on drafts of this article. We also thank David Berthelot and Tom B. Brown for help with implementation details. Finally, we thank Jakob Uszkoreit, Tao Xu, and Ashish Vaswani for helpful discussions.


\nocite{langley00}

\bibliography{reference}
\bibliographystyle{icml2019}


\end{document}